\documentclass[11pt,a4paper,logo]{lumia}

\usepackage[round,authoryear]{natbib}
\usepackage{mathtools}
\usepackage{multirow}
\usepackage{placeins}

\definecolor{bigaired}{RGB}{156,0,0}
\definecolor{color5}{HTML}{006795}
\hypersetup{
    colorlinks=true,
    linkcolor=bigaired, 
    citecolor=color5,   
    urlcolor=color5     
}
\newcommand{\method}{\textsc{P2N}}
\newcommand{\Pre}{\mathcal{F}_{\mathrm{pre}}}
\newcommand{\Core}{\mathcal{F}_{\mathrm{core}}}
\newcommand{\Suf}{\mathcal{F}_{\mathrm{suf}}}
\newcommand{\ShiftPrev}{\operatorname{ShiftPrev}}

\setheadertext{LUMIA Lab}
\correspondingemail{\emailicon~hyq718@sjtu.edu.cn, boyizeng@sjtu.edu.cn, lin.zhouhan@gmail.com \quad $^*$ Equal Contribution \quad $^\ddagger$ Corresponding Author.}
\githublink{https://github.com/LUMIA-Group/P2N}

\title{P2N: Reusing Deep Representations for Greater Effective Depth}
\author{%
Yiqin Hao$^{1,2*}$, Boyi Zeng$^{1,*}$, Zitong Wang$^{1,2,4}$, He Li$^{1}$,
Shixiang Song$^{1}$, Feichen Song$^{1}$, Yifan Liu$^{1}$, Ran Tao$^{2}$, Bryan Dai$^{2}$, Zhouhan Lin$^{1,3\ddagger}$\\
$^{1}$LUMIA Lab, School of Artificial Intelligence, Shanghai Jiao Tong University \quad$^{2}$IQuest Research \quad $^3$ Shanghai AI Laboratory \quad$^{4}$Sun Yat-sen University
}

\begin{document}

\begin{abstract}
Mathematical problem solving and autonomous agents are increasingly
important applications of language models, demanding stronger reasoning
capabilities. Computational depth is a key resource for such reasoning,
but adding model layers increases both training and inference costs,
while test-time scaling incurs substantial additional inference computation.
We introduce \method{} (Previous to Next), which reuses the previous token's deep representations—already available during generation—to increase effective computational depth with negligible decoding overhead.
Specifically, \method{} adds the previous token's deep hidden state to the
current token's shallow representation, allowing subsequent layers to build
on previously completed deep computation. This recurrence extends effective
computational depth across tokens, while introducing only a single vector
addition per decoding step. We train the recurrence using parallel
Jacobi updates restricted to a middle-layer core. Experiments with
Qwen-style models show that, at 1.5B parameters,
\method{} improves GSM8K accuracy by over 10 percentage points under both
zero-shot and few-shot prompting, with nearly unchanged decoding latency.
The benefits also extend to language modeling and average performance
on general downstream benchmarks.
\end{abstract}
\maketitle

\begin{figure}[!htbp]
    \centering
    \includegraphics[width=0.8\linewidth]{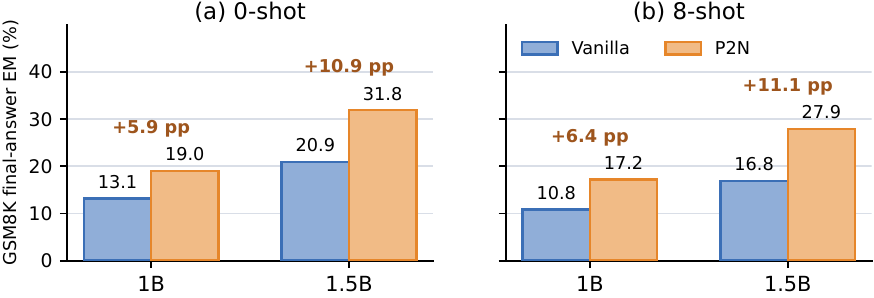}
    \caption{\textbf{Mathematical reasoning at the 1B and 1.5B scales.}
    At 1.5B, \method{} improves GSM8K accuracy over Vanilla by
    10.9 and 11.1 percentage points under zero-shot and eight-shot
    prompting, respectively, corresponding to relative gains of
    52.2\% and 65.8\%.}
    \label{fig:gsm8k}
\end{figure}

\section{Introduction}
\label{sec:introduction}

The growing use of language models for mathematical problem solving
and autonomous agents places increasing demands on their reasoning
capabilities. Computational depth is an important resource for reasoning,
enabling representations to be refined through successive stages of
computation \citep{dehghani2018universal,geiping2026scaling}. However,
adding Transformer layers increases both training and inference costs,
while test-time scaling \citep{snell2024scaling} incurs additional computation during generation.
These costs motivate a natural question: can language models obtain
greater effective depth by reusing computation that has already been
performed?

We therefore introduce \method{} (Previous to Next), which reuses
the previous token's deep representations—already available during
generation---to increase effective computational depth with negligible
decoding overhead. Specifically, for a contiguous core of Transformer
layers, \method{} adds the previous token's core output to the current
token's core input. This parameter-free feedback requires only a single
vector addition per token.

\begin{figure}[!t]
\begin{center}
\includegraphics[width=1\linewidth]{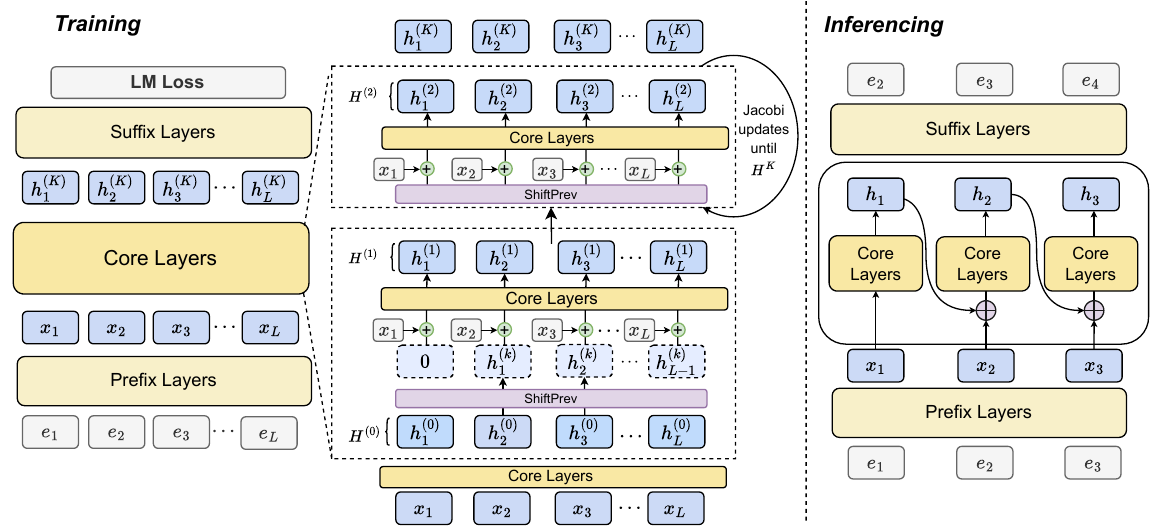}
\end{center}
\caption{\textbf{Overview of \method{}.}
The Transformer is divided into prefix, core, and suffix layers.
\textbf{Autoregressive decoding (right).} At each token, the previous token's
core output is added to the current token's prefix representation before entering
the core layers. Since the previous token's state is already available, this cross-token
feedback is naturally compatible with autoregressive decoding; however,
directly applying it during teacher-forced training would introduce a
sequential dependency across token positions.
\textbf{Parallel training (left).} The prefix processes the full
sequence once, and an initial warm pass through the core produces the starting
states. A small number of Jacobi updates then refines these states in parallel:
in each iteration, the preceding iteration's core outputs are shifted one
position to the right and added to the fixed prefix representations to form the
new inputs. Finally, the suffix layers are applied once to the final states to
compute the language-modeling loss.}
\label{fig:p2n-overview}
\end{figure}

Although this recurrence fits naturally into autoregressive decoding,
directly applying it during training would introduce a left-to-right
dependency across tokens. We address this dependency using Jacobi
iteration \citep{saad2003iterative}. After an initial warm pass, each
update combines the fixed core inputs with the preceding iteration's
outputs shifted one position to the right, allowing all token positions
to be updated in parallel. To limit the additional training cost, we
restrict the recurrent core to the middle third of the network and
evaluate the prefix and suffix only once, as illustrated in
Figure~\ref{fig:p2n-overview}.

We pretrain Qwen-style models \citep{yang2025qwen3} from 150M to 1.5B
parameters from scratch on the Pile \citep{gao2020pile}, comparing
\method{} with vanilla baselines at matched parameter and token
budgets. At 1.5B, after matched supervised fine-tuning, \method{}
improves GSM8K \citep{cobbe2021training} final-answer exact match by
10.92 and 11.07 percentage points under zero-shot and eight-shot
prompting, respectively. Decoding benchmarks show nearly unchanged
per-token latency. The benefits also extend to language modeling and
general downstream performance: \method{} consistently lowers
validation loss across scales and, at 1.5B, reduces Pile validation
perplexity from 6.72 to 6.52 while improving average zero-shot and
five-shot accuracy across nine task configurations by 1.68 and 2.11
percentage points, respectively.

\section{Related Work}
\label{sec:related-work}
\paragraph{Latent representations in large language models.} Large language models use discrete tokens at their interfaces but represent context and intermediate computation in continuous hidden states \citep{vaswani2017attention}. Prior work exploits this space by feeding hidden states back as later inputs \citep{hao2024training}, mixing text with latent tokens or continuous concepts \citep{su2025token,zhang2026soft}, and combining discrete tokens with hidden representations \citep{yue2026hybrid,amos2026latent}. Other methods shape hidden states through auxiliary pretraining objectives \citep{tack2026llm,teoh2025next}. Universal Transformer repeatedly applies shared computation across depth \citep{dehghani2018universal}, a direction later developed by Looped Transformers \citep{giannou2023looped,yang2024looped} and recurrent-depth models \citep{geiping2026scaling}. These approaches extend computation beyond discrete tokens, but may  increase training or inference cost.

\paragraph{Latent feedback in Transformers.} Some work makes deeper representations available to shallow computation by changing how historical hidden states are stored and reused. Segment-level approaches cache or compress hidden states \citep{dai2019transformer,rae2019compressive}, recurrent-memory models carry block states or memory tokens across longer contexts \citep{hutchins2022block,bulatov2022recurrent}, and feedback-based models expose higher-level history through recurrent or feedback attention \citep{fan2020addressing,ju2022staircase,hwang2024transformerfam}. Recent designs create more direct cross-token, cross-layer paths through downward connections \citep{tang2026turbo}, output-derived KVs \citep{oncescu2026recurrent}, and learned projection or fusion modules \citep{huang2026latent,cai2026t}. PonderLM-2 treats the current position’s final hidden state as an additional latent thought and feeds it back before predicting the next token \citep{zeng2025ponderlm}. The Full-bandwidth Transformer uses a learned gate to fuse the previous token's top-layer hidden state with the current token embedding \citep{wang2026full},  each feedback pass during pretraining reruns the full Transformer stack. Recirculation adds feedback after pretraining by sending deep activations back to shallow layers \citep{mozer2026recirculation}. Its feedback path is not pretrained, and decoding runs two complete model computations in parallel at each step. However, these methods either introduce additional memory, projection, or fusion components, require extra full-model computation, or add feedback only after pretraining. P2N learns a simple additive feedback connection during pretraining, preserves sequence parallelism through Jacobi updates, and reuses cached states during decoding.

\section{Method}
\label{sec:method}


We first describe \method{}  without committing to a particular subset of model
layers. Let $F$ denote the transformation performed by a causal language model,
$x_t\in\mathbb{R}^{d}$ its input representation at token position $t$, and
$h_t\in\mathbb{R}^{d}$ its output representation. For clarity, we omit the
standard causal context from the notation. A vanilla model can then be written
as:
\begin{equation}
    h_t = F(x_t).
    \label{eq:vanilla-model}
\end{equation}

\subsection{Cross-token recurrence during autoregressive decoding}
\label{sec:decode}

At autoregressive step $t$, the model receives the current input $x_t$.
Because decoding proceeds from left to right, the previous token's output
$h_{t-1}$ has already been computed. \method{} therefore defines:
\begin{equation}
    h_t = F\!\left(x_t + h_{t-1}\right),
    \qquad h_0 = 0.
    \label{eq:p2n-recurrence}
\end{equation}

\subsection{Parallel training with Jacobi iteration}
\label{sec:jacobi-training}

However, directly applying ~\autoref{eq:p2n-recurrence} during
teacher-forced training would introduce a left-to-right sequential dependency.
Although the complete input sequence is available, the state $h_{t-1}$ is an
\emph{output} of $F$ and is therefore unavailable until the preceding token has
completed its computation. A literal implementation would consequently give up the token parallelism of standard Transformer training. We avoid this serialization with a Jacobi iteration \citep{zeng2025ponderlm, saad2003iterative}.

Let $X=[x_1,\ldots,x_T]\in\mathbb{R}^{T\times d}$ denote the input sequence,
and let $H^{(k)}\in\mathbb{R}^{T\times d}$ denote the sequence of model outputs
after the $k$-th iteration. Here, the superscript $(k)$ indexes the Jacobi
iteration, whereas a subscript $t$ indexes a token position.

\paragraph{Stage 1: warm pass.}
We first process the full sequence without cross-token output feedback:
\begin{equation}
    H^{(0)} = F(X).
    \label{eq:warm-pass}
\end{equation}
The warm pass provides the initial output sequence from which the recurrent
states are constructed.

\paragraph{Stage 2: Jacobi iterations.}
Starting from $H^{(0)}$, we perform $K$ parallel updates:
\begin{equation}
    H^{(k+1)}
    = F\!\left(X + \ShiftPrev\!\left(H^{(k)}\right)\right),
    \qquad k=0,\ldots,K-1.
    \label{eq:jacobi-update}
\end{equation}
The operator $\ShiftPrev$ shifts the sequence $H^{(k)}$ one position to the
right, so that position $t$ receives the state from position $t-1$; the first
position and packed-sequence boundaries are filled with zeros. Each update
reads only the completed outputs of the preceding iteration. Therefore, all
token positions within one Jacobi iteration remain parallel, while only the
small number of iterations is sequential. The final state $H^{(K)}$ is used to
compute the language-modeling loss. All evaluations of $F$ share parameters,
and gradients propagate through the complete unrolled computation. In our
default setting, we sample $K$ uniformly from $\{2,3\}$ for each training batch \citep{zeng2025ponderlm}.

\subsection{Partial architecture}
\label{sec:partial-loop}

To reduce the cost of Jacobi training, we apply \method{} only to a contiguous
subset of Transformer layers. We partition an $L$-layer model into three
modules:
\begin{equation}
\mathcal{F}_{\mathrm{full}} = \Suf \circ \Core \circ \Pre,
\label{eq:model-partition}
\end{equation}
where the recurrent core contains $C$ layers. In
~\autoref{sec:decode} and~\autoref{sec:jacobi-training}, the transformation
$F$ is instantiated as $\Core$: only the core participates in the cross-token
recurrence and Jacobi iterations, while the prefix and suffix are each
evaluated once.

Unless stated otherwise, the prefix, core, and suffix each contain
approximately one third of the model layers, so $C\approx L/3$.
For full-network Jacobi training, the estimated training compute
relative to Vanilla is:
\begin{equation}
    \mathbb{E}[K+1]
    = 3.5,
    \qquad K\sim\mathrm{Uniform}\{2,3\}.
\end{equation}
Here, $K$ is the number of additional Jacobi iterations sampled for
each training batch, and the $+1$ accounts for the initial warm pass.
With the partial-loop design, only the $C$ core layers are repeatedly
evaluated, reducing this compute ratio to:
\begin{equation}
    \frac{(L-C)+\mathbb{E}[K+1]C}{L}
    = 1+\frac{2.5C}{L}
    \approx 1+\frac{2.5}{3}
    \approx 1.83.
\end{equation}


\providecommand{\pTwoNPlot}[2]{%
  \IfFileExists{#1}{%
    \includegraphics[width=\linewidth]{#1}%
  }{%
    \fbox{\parbox[c][#2][c]{\dimexpr\linewidth-2\fboxsep-2\fboxrule\relax}{%
      \centering\small Plot pending}}%
  }%
}

\begin{figure}[!t]
    \centering
    \includegraphics[width=\linewidth]{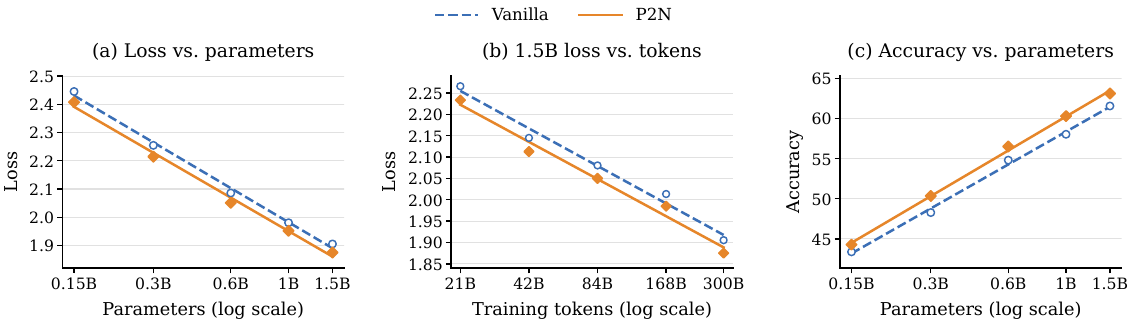}
    \caption{\textbf{Scaling behavior.}
    (a) Final Pile validation loss across parameter scales (lower is better).
    (b) Pile validation loss over training tokens at the 1.5B scale (lower is
    better). (c) Mean zero-shot accuracy across parameter scales (higher is
    better). Overall, \method{} exhibits consistently favorable behavior across
    all three scaling views.}
    \label{fig:scaling}
\end{figure}

\section{Experiments}
\label{sec:experiments}

Our experiments are organized into four parts. First, we study scaling
behavior and language modeling through large-scale pretraining. Second, we
evaluate downstream capabilities, including general benchmarks, instruction
following, and mathematical reasoning. Third, we measure autoregressive decoding
efficiency. Finally, we examine the convergence of Jacobi iteration.

\subsection{Large-Scale Pretraining}
\label{sec:large-scale-pretraining}

We pretrain a series of Qwen-style dense language models \citep{yang2025qwen3} from scratch on the Pile \citep{gao2020pile}, covering model scales of 150M, 300M, 600M, 1B, and 1.5B. We use a target
tokens-per-parameter ratio of 200 at each scale. For every model size, P2N 
and the corresponding Vanilla baseline share the same backbone architecture,
training data, token budget, and optimization settings. Detailed model
configurations and training hyperparameters are provided in
~\autoref{app:training_details}.

\subsubsection{Scaling Behavior}
\label{sec:scaling-behavior}

\autoref{fig:scaling} summarizes the scaling behavior of \method{}. Panels~(a)
and~(c) report final Pile validation loss and average downstream accuracy across
model sizes, respectively, while panel~(b) tracks Pile validation loss throughout
pretraining for the matched 1.5B models. The downstream score is the unweighted
mean of zero-shot accuracies over the seven informative task configurations
in~\autoref{tab:downstream}, excluding WinoGrande~\citep{sakaguchi2021winogrande} and ARC-Challenge~\citep{clark2018think} because
their small-model scores remain close to random guessing. All horizontal axes
use a logarithmic scale. Across all evaluated model sizes, \method{} achieves lower
validation loss and higher downstream accuracy than Vanilla, while maintaining
lower loss throughout pretraining at the 1.5B scale.


\subsubsection{Language Modeling Ability}
\label{sec:language-modeling}

\begin{figure}[!t]
    \centering
    \includegraphics[width=\linewidth]{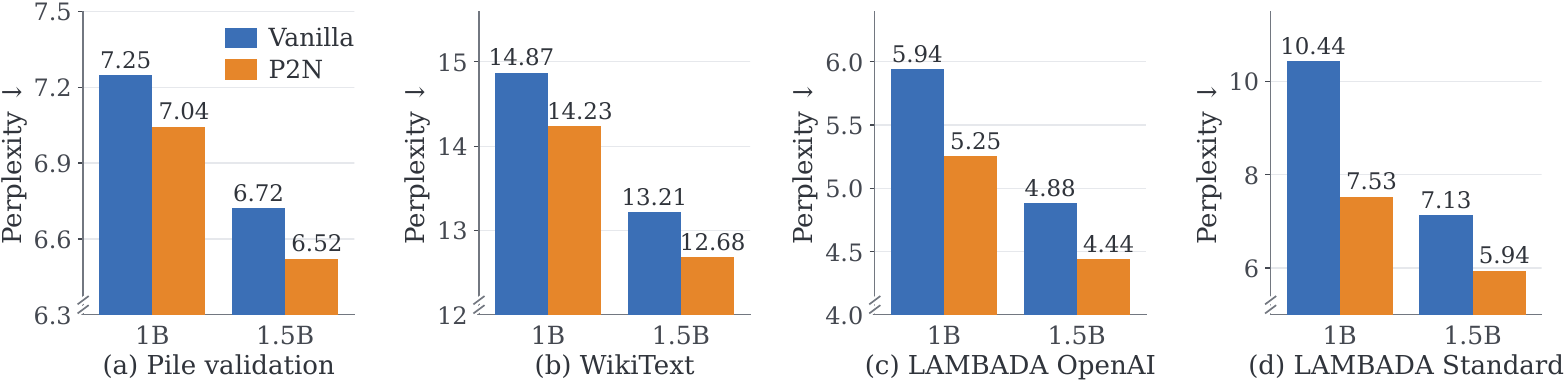}
    \caption{\textbf{Language modeling at the 1B and 1.5B scales.}
    Perplexity of Vanilla and \method{} dense models on
    (a) Pile validation, (b) WikiText-2, (c) LAMBADA OpenAI, and
    (d) LAMBADA Standard; lower is better.
    WikiText-2 uses word-level perplexity, and LAMBADA uses final-word
    perplexity. \method{} uses three Jacobi updates at evaluation time.
    Numbers above the bars denote perplexity values.
    \method{} achieves lower perplexity at both scales on every benchmark, indicating the effectiveness of this cross-token feedback method.}
    \label{fig:language-modeling}
\end{figure}

We evaluate the 1B and 1.5B dense models on Pile validation \citep{gao2020pile}, WikiText-2 \citep{merity2016pointer}, the OpenAI and Standard variants of LAMBADA \citep{paperno2016lambada}, using matched preprocessing and scoring protocols for Vanilla and \method{}.
As shown in~\autoref{fig:language-modeling}, \method{}  consistently reduces
perplexity across all four benchmarks at both model sizes.
Relative to Vanilla, the reductions are 2.8--3.0\% on Pile, 4.0--4.3\% on
WikiText-2, 9.1--11.6\% on LAMBADA OpenAI, and 16.7--27.8\% on LAMBADA
Standard. The largest relative improvements occur on the two LAMBADA
variants, indicating improved prediction of the final word from its preceding
context. These results show that the gains extend beyond the pretraining
validation set to the additional language modeling benchmarks.

\subsection{Downstream Task Evaluation}
\label{sec:downstream-evaluation}

We further assess the downstream capabilities of \method{} at the 1B and
1.5B scales through a series of experiments. These evaluations examine
whether the improvements observed during pretraining translate into
better downstream performance.

\subsubsection{General Downstream Tasks}
\label{sec:general-downstream}

We evaluate Vanilla and \method{} dense models on nine task configurations: LAMBADA-OpenAI, LAMBADA-Standard \citep{paperno2016lambada}, PIQA \citep{bisk2020piqa}, WinoGrande \citep{sakaguchi2021winogrande}, SciQ \citep{welbl2017crowdsourcing}, HellaSwag \citep{zellers2019hellaswag}, ARC-Easy, ARC-Challenge \citep{clark2018think}, and RACE \citep{lai2017race}.~\autoref{tab:downstream} reports
zero-shot and five-shot raw accuracy, together with the unweighted
average over all nine configurations. At evaluation time, \method{} uses
$K=3$ Jacobi iterations after the warm pass.

In zero-shot evaluation, \method{} improves average accuracy from
54.85\% to 56.43\% at 1B and from 57.66\% to 59.34\% at 1.5B.
The improvements also hold in the five-shot setting, where average
accuracy increases from 55.73\% to 57.42\% at 1B and from 58.20\%
to 60.31\% at 1.5B. These results demonstrate consistent gains in
overall downstream performance across both model scales and shot
settings.

\begin{table}[!t]
    \centering
    \caption{\textbf{Downstream performance at the 1B and 1.5B scales.}
    Zero-shot and five-shot raw accuracy (\%) of dense models.
    Avg. is the unweighted mean over the nine task configurations.
    Bold indicates the higher score within each matched scale and shot setting.
    \method{}  consistently improves average accuracy at both scales
    in both settings.}
    \label{tab:downstream}

    \footnotesize
    \setlength{\tabcolsep}{2.5pt}
    \renewcommand{\arraystretch}{1.18}

    \begin{tabularx}{\linewidth}{
        @{}
        c
        l
        *{10}{>{\centering\arraybackslash}X}
        @{}
    }
        \toprule

        \multirow{2}{*}{Scale}
        & \multirow{2}{*}{Method}
        & \multicolumn{2}{c}{LAMBADA}
        & \multirow{2}{*}{PIQA}
        & \multirow{2}{*}{\shortstack{Wino\\Grande}}
        & \multirow{2}{*}{SciQ}
        & \multirow{2}{*}{\shortstack{Hella\\Swag}}
        & \multicolumn{2}{c}{ARC}
        & \multirow{2}{*}{RACE}
        & \multirow{2}{*}{Avg.} \\

        \cmidrule(lr){3-4}
        \cmidrule(lr){9-10}

        &
        & OpenAI
        & Standard
        &
        &
        &
        &
        & Easy
        & Challenge
        &
        & \\

        \midrule

        \multicolumn{12}{c}{\textbf{Zero-shot}} \\
        \midrule

        1B & Vanilla
        & 61.73
        & 49.91
        & 71.06
        & \textbf{59.91}
        & 88.70
        & 41.62
        & 60.02
        & \textbf{27.39}
        & 33.30
        & 54.85 \\

        1B & \method{}
        & \textbf{64.22}
        & \textbf{57.13}
        & \textbf{72.31}
        & 58.64
        & \textbf{89.50}
        & \textbf{43.45}
        & \textbf{61.45}
        & 27.05
        & \textbf{34.16}
        & \textbf{56.43} \\

        \addlinespace[4pt]

        1.5B & Vanilla
        & 66.23
        & 57.36
        & \textbf{74.16}
        & 59.91
        & 88.60
        & 45.05
        & 64.14
        & 27.99
        & 35.50
        & 57.66 \\

        1.5B & \method{}
        & \textbf{67.92}
        & \textbf{61.46}
        & 73.18
        & \textbf{62.27}
        & \textbf{90.70}
        & \textbf{46.74}
        & \textbf{64.90}
        & \textbf{29.86}
        & \textbf{37.03}
        & \textbf{59.34} \\

        \midrule

        \multicolumn{12}{c}{\textbf{Five-shot}} \\
        \midrule

        1B & Vanilla
        & 57.62
        & 49.18
        & \textbf{72.58}
        & 60.06
        & 93.40
        & 41.68
        & 63.38
        & 29.10
        & 34.55
        & 55.73 \\

        1B & \method{}
        & \textbf{59.21}
        & \textbf{56.47}
        & 72.36
        & \textbf{61.09}
        & \textbf{94.10}
        & \textbf{43.61}
        & \textbf{64.77}
        & \textbf{29.27}
        & \textbf{35.89}
        & \textbf{57.42} \\

        \addlinespace[4pt]

        1.5B & Vanilla
        & 61.27
        & 52.63
        & \textbf{74.48}
        & 61.80
        & \textbf{94.20}
        & 45.18
        & 66.33
        & 32.59
        & 35.31
        & 58.20 \\

        1.5B & \method{}
        & \textbf{62.90}
        & \textbf{60.88}
        & 74.05
        & \textbf{65.43}
        & 93.70
        & \textbf{47.13}
        & \textbf{68.18}
        & \textbf{33.62}
        & \textbf{36.94}
        & \textbf{60.31} \\

        \bottomrule
    \end{tabularx}
\end{table}

\subsubsection{Instruction-Following Ability}
\label{sec:instruction-following}

We fine-tune the 1B and 1.5B dense models and their vanilla counterparts on Alpaca \citep{taori2023stanford}, and evaluate their instruction-following ability on MT-Bench \citep{zheng2023judging}.
Fine-tuning, generation, and judging settings are matched within each
model-scale comparison. We use GPT-5.5 as the judge and report scores
averaged over both conversation turns.

\begin{figure}[!t]
    \centering
    \includegraphics[width=\linewidth]{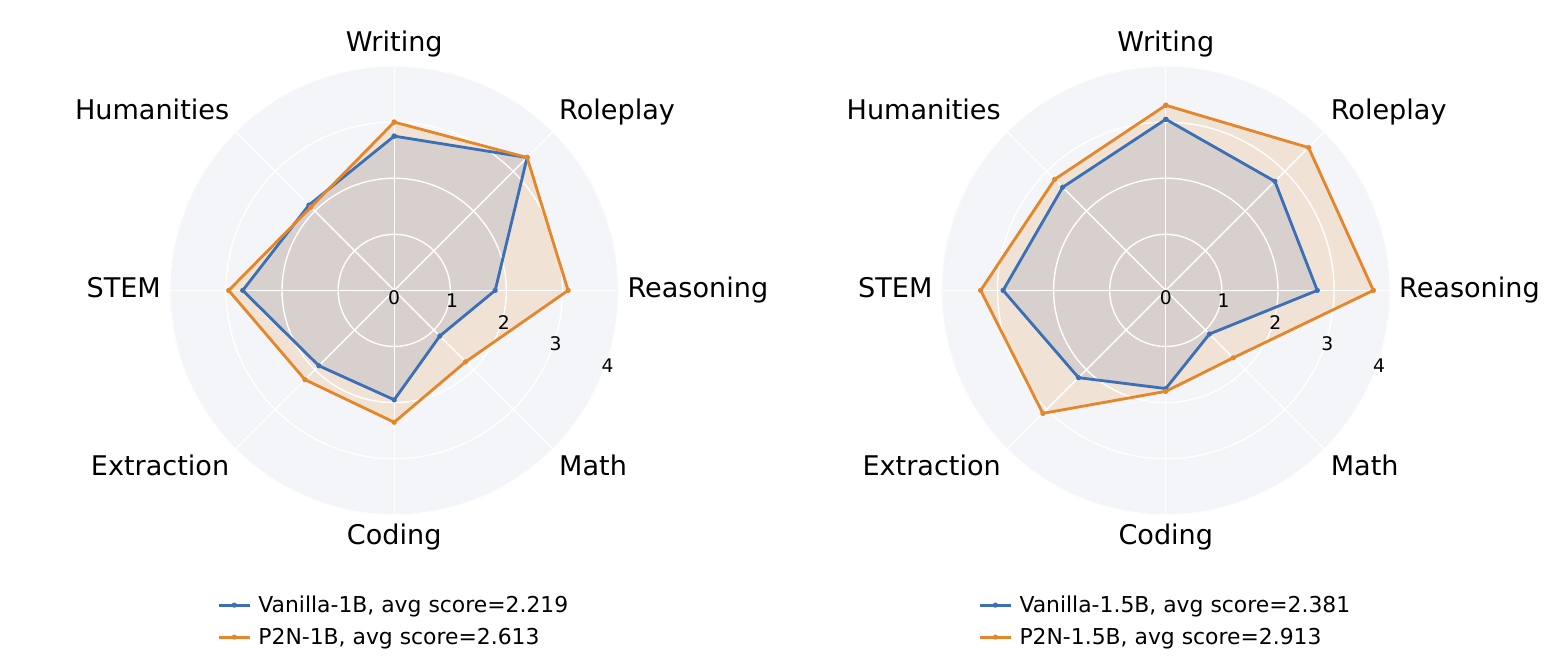}
    \caption{\textbf{Instruction-following ability.}
    Category-wise MT-Bench scores for 1B (left) and 1.5B (right)
    dense models after Alpaca fine-tuning, evaluated by GPT-5.5.
    Each category score averages both conversation turns;
    legends report the overall mean across all 80 questions
    and both turns.
    \method{} improves the overall score at both scales,
    with gains across all eight categories at 1.5B.
    Both panels use the same radial scale.}
    \label{fig:mtbench}
\end{figure}

As shown in~\autoref{fig:mtbench}, \method{} improves the overall
MT-Bench score from 2.219 to 2.613 at 1B and from 2.381 to 2.913 at
1.5B, yielding gains of 0.394 and 0.532 points, respectively.
At 1B, \method{} improves six of the eight categories, matches the
vanilla model on roleplay, and slightly trails it on humanities.
At 1.5B, improvements extend to all eight categories, with the largest
gains in reasoning, extraction, and roleplay.
These results suggest that the benefits of \method{} transfer to
instruction-following after supervised fine-tuning.

\subsubsection{Mathematical Reasoning}
\label{sec:mathematical-reasoning}

We fine-tune the pretrained 1B and 1.5B dense models on the official GSM8K training split \citep{cobbe2021training} and evaluate final-answer exact match (EM) on all 1,319 test examples under zero-shot and eight-shot prompting.
Fine-tuning budgets are matched within each scale, and prompting
formats, decoding settings, and answer-extraction procedures are
matched between Vanilla and \method{} within each evaluation setting.

As shown in~\autoref{fig:gsm8k}, \method{} improves zero-shot EM by
5.91 and 10.92 percentage points at 1B and 1.5B, respectively.
The improvements also hold under eight-shot prompting, with gains
of 6.37 and 11.07 percentage points at the corresponding scales.
These results demonstrate improved mathematical reasoning performance
across both model scales and prompting settings.

\subsection{Decoding Efficiency}
\label{sec:decoding-efficiency}
During autoregressive decoding, \method{} introduces only one additional vector
addition per token, without repeating any Transformer layer. We benchmark the
1.5B dense model on a single GPU using BF16, SDPA, and the standard KV
cache. Each run uses a 128-token prompt and measures 2,048 decode steps after
16 untimed warm-up steps. We report averages over five runs for each batch size.

\autoref{tab:decoding-efficiency} reports time per output token (TPOT),
aggregate decoding throughput, and relative latency. Across batch sizes,
\method{} consistently maintains nearly identical latency to Vanilla,
indicating negligible additional decoding overhead.

\begin{table}[!t]
    \centering
    \caption{\textbf{Autoregressive decoding efficiency of the 1.5B dense
    model on a single GPU.}
    TPOT denotes time per output token, and throughput counts generated
    tokens per second across the batch. Results are averaged over five
    paired runs generating 2,048 tokens per sequence.
    Rel. latency is the ratio of \method{} TPOT to Vanilla TPOT.
    \method{} maintains near-vanilla latency across batch sizes.}
    \label{tab:decoding-efficiency}

    \small
    \begin{tabular}{rccccc}
        \toprule
        Batch
        & \multicolumn{2}{c}{TPOT (ms)}
        & \multicolumn{2}{c}{Throughput (tok/s)}
        & Rel. latency \\
        \cmidrule(lr){2-3}
        \cmidrule(lr){4-5}
        size & Vanilla & \method{}
             & Vanilla & \method{}
             & \method{} / Vanilla \\
        \midrule
        1  & 16.442 & 16.494 &   60.8 &   60.6 & $1.003\times$ \\
        4  & 16.983 & 16.991 &  235.5 &  235.4 & $1.000\times$ \\
        16 & 17.079 & 17.143 &  936.8 &  933.3 & $1.004\times$ \\
        32 & 22.298 & 22.352 & 1435.1 & 1431.7 & $1.002\times$ \\
        \bottomrule
    \end{tabular}
\end{table}

\subsection{Convergence Diagnostics of Jacobi Iteration}
\label{sec:jacobi-diagnostics}

We further examine the convergence behavior of Jacobi iteration on the
1.5B \method{} model. Let
$H^{(k)} \in \mathbb{R}^{T\times d}$
denote the core states after $k$ Jacobi iterations, with $H^{(0)}$
obtained from the warm pass. We also denote by
$H_{\mathrm{seq}} \in \mathbb{R}^{T\times d}$
the corresponding core states obtained by sequentially applying the same
recurrence from left to right. We measure convergence using the RMSE between
consecutive Jacobi states and the RMSE between Jacobi and sequential states:
\begin{equation}
    \Delta_k
    =
    \frac{
        \lVert H^{(k)} - H^{(k-1)} \rVert_{\mathrm{F}}
    }{\sqrt{Td}},
    \qquad
    d_k
    =
    \frac{
        \lVert H^{(k)} - H_{\mathrm{seq}} \rVert_{\mathrm{F}}
    }{\sqrt{Td}}.
    \label{eq:jacobi-diagnostics}
\end{equation}

\begin{figure}[!t]
    \centering
    \includegraphics[width=\linewidth]{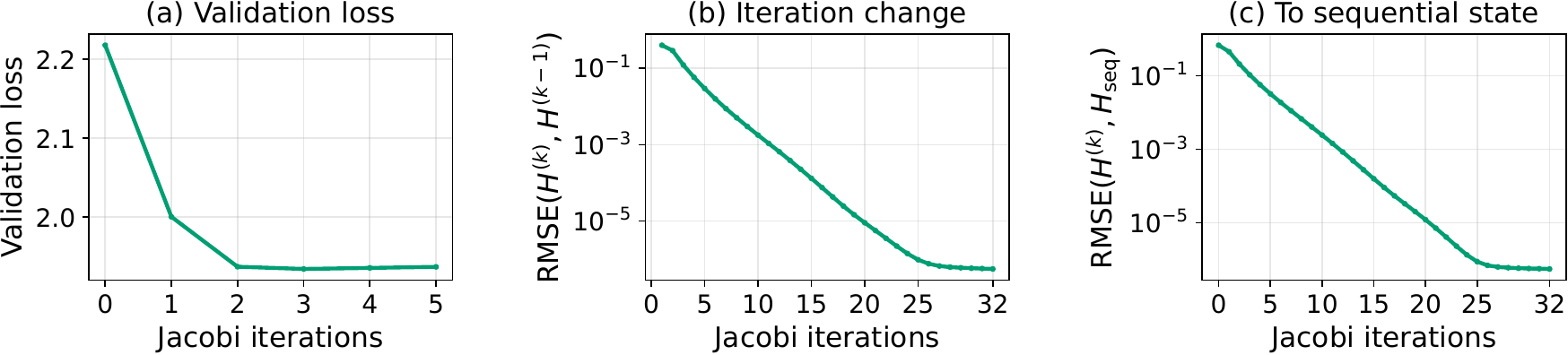}
    \caption{\textbf{Convergence diagnostics of Jacobi iteration on the
    1.5B \method{} model.}
    \textbf{(a)} Validation loss on 4096 eval samples under different numbers of Jacobi iterations.
    \textbf{(b)} RMSE between consecutive core states,
    $\Delta_k =
    \lVert H^{(k)} - H^{(k-1)} \rVert_{\mathrm{F}} / \sqrt{Td}$.
    \textbf{(c)} RMSE between Jacobi and sequential core states,
    $d_k =
    \lVert H^{(k)} - H_{\mathrm{seq}} \rVert_{\mathrm{F}} / \sqrt{Td}$.
    Validation loss quickly stabilizes, while both state errors continue
    decreasing toward zero as additional iterations are applied.}
    \label{fig:jacobi-diagnostics}
\end{figure}

\autoref{fig:jacobi-diagnostics} shows clear convergence as the number of
Jacobi iterations increases. Validation loss decreases from 2.2180 at $k=0$
to 1.9339 at $k=3$ and remains nearly unchanged afterwards, reaching 1.9365
at $k=5$. Meanwhile, both $\Delta_k$ and $d_k$ continue to decrease toward
zero: they fall below $10^{-4}$ by $k=16$ and reach approximately
$5.5\times10^{-7}$ at $k=32$. These results show that the Jacobi states
progressively converge to a stable solution that closely matches sequential
inference. Importantly, although \method{} is trained using only random
$K\in\{2,3\}$ Jacobi iterations, this training scheme is sufficient to learn
stable and convergent iterative dynamics.

\section{Ablation Studies}
\label{sec:ablations}

We conduct a series of ablations on the 21-layer, 600M dense model trained on 120B tokens
from the Pile. Unless otherwise specified, we use direct sum and a
middle-third recurrent core, with $K$ sampled uniformly from $\{2,3\}$
for each training batch. Here, $K$ counts Jacobi iterations after the
warm pass. Each ablation changes only the factor under study.

\subsection{Feedback Formulation}
\label{sec:ablation-feedback}

We first examine how to combine the current core input $x_t$ with the
previous token's core output $h_{t-1}$, considering residual-scale
treatment and learnable mixing.

\paragraph{Residual-scale treatment.}
Our default formulation directly uses $x_t+h_{t-1}$ as core input.
To examine whether explicit residual-scale control is beneficial, we
compare it with \emph{RMS matching}, which rescales the sum to match
the original input's RMS, and \emph{compute-only},
which uses $x_t+h_{t-1}$ only as the attention pre-normalization input
in the core block while retaining $x_t$ on the residual shortcut.

\paragraph{Learnable mixing.}
We also compare direct sum with channel-wise and scalar gates that
learn how strongly to weight the feedback. The channel-wise variant
uses $x_t+g\odot h_{t-1}$, where $g\in\mathbb{R}^{d}$ is initialized
to $\mathbf{1}$. The scalar variant instead learns a single feedback
weight shared across channels.

\autoref{tab:ablation-feedback} shows that RMS matching and compute-only
increase validation loss by 0.0032 and 0.0154, respectively, relative
to direct sum. Channel-wise and scalar gates provide only small gains,
reducing loss by 0.0020 and 0.0017. Thus, neither explicit scale treatment
nor learnable mixing offers a compelling advantage over the simpler
formulation, and we retain direct sum as our default.

\begin{table}[!htbp]
    \centering
    \caption{\textbf{Feedback formulation.}
    Validation loss on 600M model trained for 120B tokens to ablate residual-scale treatments (a) and learnable
    mixing strategies (b). Loss gaps are measured relative to direct
    sum; negative values indicate improvement. Scale treatments
    increase loss, while learnable gates offer only small gains.}
    \label{tab:ablation-feedback}

    \small
    \setlength{\tabcolsep}{4pt}
    \renewcommand{\arraystretch}{1.12}
    \begin{tabular}{@{}lrr@{\hspace{1.8em}}lrr@{}}
        \toprule
        \multicolumn{3}{c}{\textbf{(a) Residual-scale treatment}}
        & \multicolumn{3}{c}{\textbf{(b) Learnable mixing}} \\
        \cmidrule(r){1-3}
        \cmidrule(l){4-6}
        Method & Val. loss $\downarrow$ & Loss gap
        & Method & Val. loss $\downarrow$ & Loss gap \\
        \midrule
        Direct sum (\method{}) & \textbf{2.0511} & $0.0000$
        & Direct sum (\method{}) & 2.0511 & $0.0000$ \\
        \addlinespace[3pt]
        RMS matching & 2.0543 & $+0.0032$
        & Channel-wise gate & \textbf{2.0491} & $-0.0020$ \\
        Compute-only & 2.0665 & $+0.0154$
        & Scalar gate & 2.0494 & $-0.0017$ \\
        \bottomrule
    \end{tabular}
\end{table}

\subsection{Recurrent Core Depth}
\label{sec:ablation-core-depth}

We next vary the recurrent core depth to examine the trade-off between
model quality and training compute. We compare Vanilla with recurrence
over the middle 7 layers, the middle 11 layers (approximately half
the network), and all 21 layers.

As shown in~\autoref{tab:ablation-core-depth}, the middle-third core
reduces validation loss from 2.0859 to 2.0511. Extending the core to
11 layers provides only a further 0.0058 reduction while requiring
approximately 26\% more estimated training compute. Full-network
recurrence is the most expensive configuration yet performs worse
than Vanilla. The middle-third core therefore offers a favorable
quality--compute trade-off, capturing most of the improvement achieved
by the 11-layer core at a lower training cost.

\begin{table}[!t]
    \centering
    \caption{\textbf{Recurrent core depth.}
    Validation loss and estimated training compute relative to Vanilla on 600M model trained for 120B tokens.
    Compute ratios use the layer-based approximation $1+2.5C/21$,
    where $C$ is the number of recurrent layers.
    The middle-third core provides most of the gain from the 11-layer
    core at a lower estimated compute cost.}
    \label{tab:ablation-core-depth}

    \small
    \setlength{\tabcolsep}{8pt}
    \renewcommand{\arraystretch}{1.12}
    \begin{tabular}{@{}lccc@{}}
        \toprule
        Method & Core layers & Relative compute
        & Val. loss $\downarrow$ \\
        \midrule
        Vanilla
        & 0 & $1.00\times$ & 2.0859 \\
        \midrule
        Middle $1/3$ (\method{})
        & 7 & $1.83\times$ & 2.0511 \\
        Middle ${\approx}1/2$
        & 11 & $2.31\times$ & \textbf{2.0453} \\
        Full
        & 21 & $3.50\times$ & 2.1046 \\
        \bottomrule
    \end{tabular}
\end{table}

\subsection{Recurrent Core Placement}
\label{sec:ablation-core-placement}

With the core depth fixed at seven layers, we then compare placement
in the first, middle, and last thirds of the network. This isolates
the effect of placement under the same recurrent-compute budget.

\autoref{tab:ablation-core-placement} shows that the middle-third
placement achieves the lowest validation loss. Placing the core in
the first third yields only a small improvement over Vanilla,
whereas placing it in the last third performs worse than Vanilla.
Together with the depth ablation, these results support applying
recurrence to the middle third of the network.

\begin{figure}[!t]
    \centering
    \includegraphics[width=0.9\linewidth]{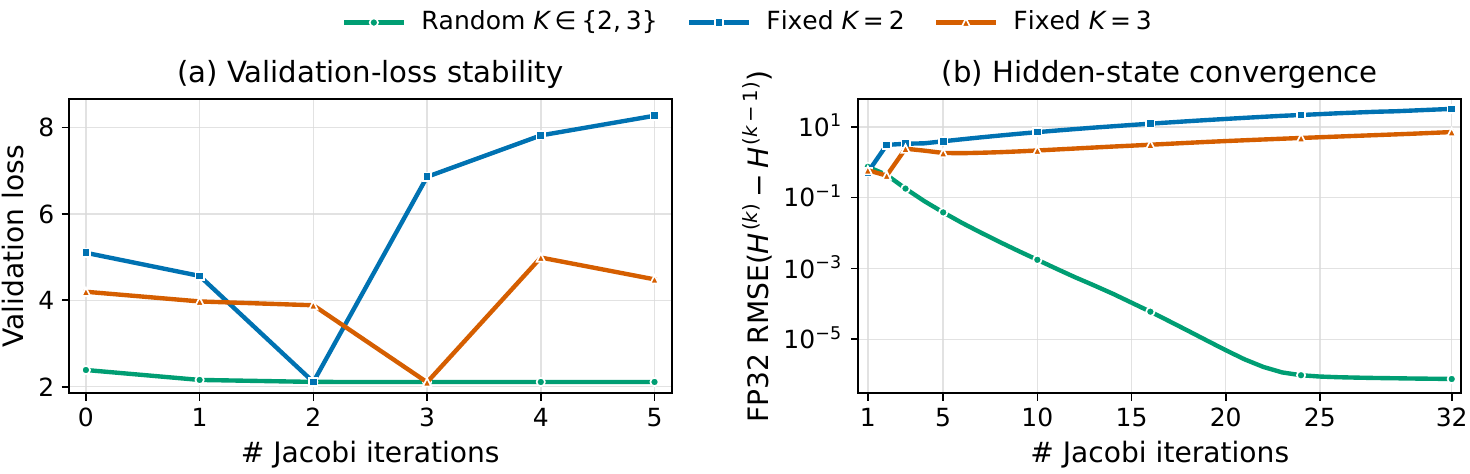}
    \caption{\textbf{Effect of sampling the number of Jacobi iterations.}
    \textbf{(a)} Validation loss over $k=0,\ldots,5$ evaluation-time
    Jacobi iterations.
    \textbf{(b)} RMSE between consecutive core states, $\Delta_k$,
    over $k=1,\ldots,32$ iterations.
    Random $K\in\{2,3\}$ yields stable loss and state differences
    approaching zero, whereas fixed-$K$ models deteriorate beyond
    their training iteration count.}
    \label{fig:ablation-jacobi}
\end{figure}

\begin{table}[!htbp]
    \centering
    \caption{\textbf{Recurrent core placement.}
    Validation loss and loss gaps relative to Vanilla on 600M model trained for 120B tokens to ablate different
    seven-layer core placements. Among recurrent configurations
    with the same compute budget, the middle-third placement
    achieves the lowest loss.}
    \label{tab:ablation-core-placement}

    \small
    \setlength{\tabcolsep}{8pt}
    \renewcommand{\arraystretch}{1.12}
    \begin{tabular}{@{}lrr@{}}
        \toprule
        Core placement & Val. loss $\downarrow$ & Loss gap \\
        \midrule
        Vanilla (no recurrence) & 2.0859 & $0.0000$ \\
        \midrule
        First $1/3$             & 2.0792 & $-0.0067$ \\
        Middle $1/3$ (\method{}) & \textbf{2.0511} & $-0.0348$ \\
        Last $1/3$              & 2.0969 & $+0.0110$ \\
        \bottomrule
    \end{tabular}
\end{table}

\subsection{Sampling Jacobi Iterations}
\label{sec:ablation-jacobi}

We compare training with fixed $K=2$ or $K=3$ Jacobi iterations against
our default random $K\in\{2,3\}$ schedule. Following the convergence
analysis in~\autoref{sec:jacobi-diagnostics}, we examine validation
loss and the RMSE between consecutive core states, $\Delta_k$, as the
number of evaluation-time Jacobi iterations increases.

As shown in~\autoref{fig:ablation-jacobi}, random sampling yields
nearly stable validation loss while $\Delta_k$ decreases toward zero.
In contrast, fixed-$K$ models specialize to their training iteration
count: beyond that count, validation loss rises sharply and $\Delta_k$
grows, indicating that the hidden states do not stabilize within the
evaluated range. These results support randomizing the number of
training-time Jacobi iterations to avoid depth specialization and
promote stable convergence.

\section{Conclusion}
\label{sec:conclusion}

We introduced \method{}, a parameter-free latent feedback mechanism
that allows each token to build on the previous token's deep computation
by adding its core output to the current core input. Jacobi iterations
preserve token parallelism during training, while restricting recurrence
to the middle third of the network limits the additional training cost.
Experiments on Qwen-style dense models from 150M to 1.5B parameters
demonstrate consistent improvements in language modeling, with gains
extending to downstream tasks and mathematical reasoning after
supervised fine-tuning. Architectural ablations support our choice of
direct addition and a middle-third core, while convergence diagnostics
show that randomized Jacobi training promotes stable iterative behavior
beyond the training iteration counts. Decoding benchmarks further show
that these improvements come with negligible additional per-token
latency. Together, these findings highlight the value of reusing
previously completed deep computation across tokens to improve language
model capabilities while maintaining efficient autoregressive decoding.

\section*{Acknowledgement}
This work is sponsored by the National Natural Science Foundation of China (NSFC) grant (No. 62576211) and the Specialized Program on Fundamental Research from Science and Technology Commission of Shanghai Municipality (No. 2025SHZDZX025G09).


\clearpage
\bibliography{references}

@article{vaswani2017attention,
  title={Attention is all you need},
  author={Vaswani, Ashish and Shazeer, Noam and Parmar, Niki and Uszkoreit, Jakob and Jones, Llion and Gomez, Aidan N and Kaiser, {\L}ukasz and Polosukhin, Illia},
  journal={Advances in neural information processing systems},
  volume={30},
  year={2017}
}

@article{hao2024training,
  title={Training large language models to reason in a continuous latent space},
  author={Hao, Shibo and Sukhbaatar, Sainbayar and Su, DiJia and Li, Xian and Hu, Zhiting and Weston, Jason and Tian, Yuandong},
  journal={arXiv preprint arXiv:2412.06769},
  year={2024}
}

@article{su2025token,
  title={Token assorted: Mixing latent and text tokens for improved language model reasoning},
  author={Su, DiJia and Zhu, Hanlin and Xu, Yingchen and Jiao, Jiantao and Tian, Yuandong and Zheng, Qinqing},
  journal={arXiv preprint arXiv:2502.03275},
  year={2025}
}

@article{zhang2026soft,
  title={Soft thinking: Unlocking the reasoning potential of llms in continuous concept space},
  author={Zhang, Zhen and He, Xuehai and Yan, Weixiang and Shen, Ao and Zhao, Chenyang and Wang, Xin},
  journal={Advances in Neural Information Processing Systems},
  volume={38},
  pages={168990--169012},
  year={2026}
}

@article{yue2026hybrid,
  title={Hybrid latent reasoning via reinforcement learning},
  author={Yue, Zhenrui and Jin, Bowen and Zeng, Huimin and Zhuang, Honglei and Qin, Zhen and Yoon, Jinsung and Shang, Lanyu and Han, Jiawei and Wang, Dong},
  journal={Advances in Neural Information Processing Systems},
  volume={38},
  pages={5501--5530},
  year={2026}
}

@article{amos2026latent,
  title={Latent reasoning with supervised thinking states},
  author={Amos, Ido and Caciularu, Avi and Geva, Mor and Globerson, Amir and Herzig, Jonathan and Shani, Lior and Szpektor, Idan},
  journal={arXiv preprint arXiv:2602.08332},
  year={2026}
}

@article{wang2026full,
  title={Full-bandwidth transformer},
  author={Wang, Xi and Cai, Ziyang and Zhan, Zheng and Dong, Harry and Fan, Ying and de Rosa, Gustavo and Pearce, Tim and Langford, John},
  journal={arXiv preprint arXiv:2608.08888},
  year={2026}
}

@inproceedings{tack2026llm,
  title={Llm pretraining with continuous concepts},
  author={Tack, Jihoon and Lanchantin, Jack and Dwivedi-Yu, Jane and Cohen, Andrew and Kulikov, Ilia and Lan, Janice and Hao, Shibo and Tian, Yuandong and Weston, Jason E and Li, Xian},
  booktitle={International Conference on Learning Representations},
  volume={2026},
  pages={54214--54231},
  year={2026}
}

@article{teoh2025next,
  title={Next-latent prediction transformers learn compact world models},
  author={Teoh, Jayden and Tomar, Manan and Ahn, Kwangjun and Hu, Edward S and Pearce, Tim and Sharma, Pratyusha and Krishnamurthy, Akshay and Islam, Riashat and Lamb, Alex and Langford, John},
  journal={arXiv preprint arXiv:2511.05963},
  year={2025}
}

@article{dehghani2018universal,
  title={Universal transformers},
  author={Dehghani, Mostafa and Gouws, Stephan and Vinyals, Oriol and Uszkoreit, Jakob and Kaiser, {\L}ukasz},
  journal={arXiv preprint arXiv:1807.03819},
  year={2018}
}

@inproceedings{giannou2023looped,
  title={Looped transformers as programmable computers},
  author={Giannou, Angeliki and Rajput, Shashank and Sohn, Jy-yong and Lee, Kangwook and Lee, Jason D and Papailiopoulos, Dimitris},
  booktitle={International Conference on Machine Learning},
  pages={11398--11442},
  year={2023},
  organization={PMLR}
}

@inproceedings{yang2024looped,
  title={Looped transformers are better at learning learning algorithms},
  author={Yang, Liu and Lee, Kangwook and Nowak, Robert and Papailiopoulos, Dimitris},
  booktitle={International conference on learning representations},
  volume={2024},
  pages={42195--42214},
  year={2024}
}

@article{geiping2026scaling,
  title={Scaling up test-time compute with latent reasoning: A recurrent depth approach},
  author={Geiping, Jonas and McLeish, Sean and Jain, Neel and Kirchenbauer, John and Singh, Siddharth and Bartoldson, Brian and Kailkhura, Bhavya and Bhatele, Abhinav and Goldstein, Tom},
  journal={Advances in Neural Information Processing Systems},
  volume={38},
  pages={41340--41391},
  year={2026}
}

@inproceedings{dai2019transformer,
  title={Transformer-xl: Attentive language models beyond a fixed-length context},
  author={Dai, Zihang and Yang, Zhilin and Yang, Yiming and Carbonell, Jaime G and Le, Quoc and Salakhutdinov, Ruslan},
  booktitle={Proceedings of the 57th annual meeting of the association for computational linguistics},
  pages={2978--2988},
  year={2019}
}

@article{rae2019compressive,
  title={Compressive transformers for long-range sequence modelling},
  author={Rae, Jack W and Potapenko, Anna and Jayakumar, Siddhant M and Lillicrap, Timothy P},
  journal={arXiv preprint arXiv:1911.05507},
  year={2019}
}

@article{fan2020addressing,
  title={Addressing some limitations of transformers with feedback memory},
  author={Fan, Angela and Lavril, Thibaut and Grave, Edouard and Joulin, Armand and Sukhbaatar, Sainbayar},
  journal={arXiv preprint arXiv:2002.09402},
  year={2020}
}

@article{ju2022staircase,
  title={Staircase attention for recurrent processing of sequences},
  author={Ju, Da and Roller, Stephen and Sukhbaatar, Sainbayar and Weston, Jason E},
  journal={Advances in neural information processing systems},
  volume={35},
  pages={13203--13213},
  year={2022}
}

@article{hutchins2022block,
  title={Block-recurrent transformers},
  author={Hutchins, DeLesley and Schlag, Imanol and Wu, Yuhuai and Dyer, Ethan and Neyshabur, Behnam},
  journal={Advances in neural information processing systems},
  volume={35},
  pages={33248--33261},
  year={2022}
}

@article{bulatov2022recurrent,
  title={Recurrent memory transformer},
  author={Bulatov, Aydar and Kuratov, Yury and Burtsev, Mikhail},
  journal={Advances in neural information processing systems},
  volume={35},
  pages={11079--11091},
  year={2022}
}

@article{hwang2024transformerfam,
  title={Transformerfam: Feedback attention is working memory},
  author={Hwang, Dongseong and Wang, Weiran and Huo, Zhuoyuan and Sim, Khe Chai and Mengibar, Pedro Moreno},
  journal={arXiv preprint arXiv:2404.09173},
  year={2024}
}

@article{tang2026turbo,
  title={Turbo Connection: Reasoning as Information Flow from Higher to Lower Layers},
  author={Tang, Mohan and Lu, Sidi},
  journal={arXiv preprint arXiv:2602.17993},
  year={2026}
}

@article{oncescu2026recurrent,
  title={The recurrent transformer: Greater effective depth and efficient decoding},
  author={Oncescu, Costin-Andrei and Morwani, Depen and Jelassi, Samy and Meterez, Alexandru and Kwun, Mujin and Kakade, Sham},
  journal={arXiv preprint arXiv:2604.21215},
  year={2026}
}

@article{huang2026latent,
  title={Latent Recurrent Transformer: Architecture Exploration, Training Strategies, and Scaling Behavior},
  author={Huang, Zeyi and He, Xuehai and Ren, LiLiang and Wang, Yiping and Peng, Baolin and Cheng, Hao and Wang, Shuohang and He, Pengcheng and Gao, Jianfeng and Lee, Yong Jae and others},
  journal={arXiv preprint arXiv:2605.26797},
  year={2026}
}

@article{cai2026t,
  title={T\^{} 2MLR: Transformer with Temporal Middle-Layer Recurrence},
  author={Cai, Ziyang and Zhu, Xingyu and Dong, Yihe and He, Yinghui and Arora, Sanjeev},
  journal={arXiv preprint arXiv:2607.15178},
  year={2026}
}

@article{zeng2025ponderlm,
  title={Ponderlm-2: Pretraining llm with latent thoughts in continuous space},
  author={Zeng, Boyi and Li, He and Song, Shixiang and Wang, Yixuan and Wang, Zitong and He, Ziwei and Wang, Xinbing and Lin, Zhouhan},
  journal={arXiv preprint arXiv:2509.23184},
  year={2025}
}

@article{mozer2026recirculation,
  title={Recirculation},
  author={Mozer, Michael C and Siddiqui, Shoaib Ahmed and Sawyer, Danny and Sanyal, Sunny and Liu, Rosanne},
  journal={arXiv preprint arXiv:2608.17981},
  year={2026}
}

@article{gao2020pile,
  title={The pile: An 800gb dataset of diverse text for language modeling},
  author={Gao, Leo and Biderman, Stella and Black, Sid and Golding, Laurence and Hoppe, Travis and Foster, Charles and Phang, Jason and He, Horace and Thite, Anish and Nabeshima, Noa and others},
  journal={arXiv preprint arXiv:2101.00027},
  year={2020}
}

@article{yang2025qwen3,
  title={Qwen3 technical report},
  author={Yang, An and Li, Anfeng and Yang, Baosong and Zhang, Beichen and Hui, Binyuan and Zheng, Bo and Yu, Bowen and Gao, Chang and Huang, Chengen and Lv, Chenxu and others},
  journal={arXiv preprint arXiv:2505.09388},
  year={2025}
}

@article{merity2016pointer,
  title={Pointer sentinel mixture models},
  author={Merity, Stephen and Xiong, Caiming and Bradbury, James and Socher, Richard},
  journal={arXiv preprint arXiv:1609.07843},
  year={2016}
}

@inproceedings{paperno2016lambada,
  title={The LAMBADA dataset: Word prediction requiring a broad discourse context},
  author={Paperno, Denis and Kruszewski, Germ{\'a}n and Lazaridou, Angeliki and Pham, Ngoc-Quan and Bernardi, Raffaella and Pezzelle, Sandro and Baroni, Marco and Boleda, Gemma and Fern{\'a}ndez, Raquel},
  booktitle={Proceedings of the 54th annual meeting of the association for computational linguistics (volume 1: Long papers)},
  pages={1525--1534},
  year={2016}
}

@inproceedings{bisk2020piqa,
  title={Piqa: Reasoning about physical commonsense in natural language},
  author={Bisk, Yonatan and Zellers, Rowan and Gao, Jianfeng and Choi, Yejin and others},
  booktitle={Proceedings of the AAAI conference on artificial intelligence},
  volume={34},
  number={05},
  pages={7432--7439},
  year={2020}
}

@article{sakaguchi2021winogrande,
  title={Winogrande: An adversarial winograd schema challenge at scale},
  author={Sakaguchi, Keisuke and Bras, Ronan Le and Bhagavatula, Chandra and Choi, Yejin},
  journal={Communications of the ACM},
  volume={64},
  number={9},
  pages={99--106},
  year={2021},
  publisher={ACM New York, NY, USA}
}

@inproceedings{welbl2017crowdsourcing,
  title={Crowdsourcing multiple choice science questions},
  author={Welbl, Johannes and Liu, Nelson F and Gardner, Matt},
  booktitle={Proceedings of the 3rd Workshop on Noisy User-generated Text},
  pages={94--106},
  year={2017}
}

@inproceedings{zellers2019hellaswag,
  title={Hellaswag: Can a machine really finish your sentence?},
  author={Zellers, Rowan and Holtzman, Ari and Bisk, Yonatan and Farhadi, Ali and Choi, Yejin},
  booktitle={Proceedings of the 57th annual meeting of the association for computational linguistics},
  pages={4791--4800},
  year={2019}
}

@article{clark2018think,
  title={Think you have solved question answering? try arc, the ai2 reasoning challenge},
  author={Clark, Peter and Cowhey, Isaac and Etzioni, Oren and Khot, Tushar and Sabharwal, Ashish and Schoenick, Carissa and Tafjord, Oyvind},
  journal={arXiv preprint arXiv:1803.05457},
  year={2018}
}

@inproceedings{lai2017race,
  title={Race: Large-scale reading comprehension dataset from examinations},
  author={Lai, Guokun and Xie, Qizhe and Liu, Hanxiao and Yang, Yiming and Hovy, Eduard},
  booktitle={Proceedings of the 2017 conference on empirical methods in natural language processing},
  pages={785--794},
  year={2017}
}

@misc{taori2023stanford,
  title={Stanford alpaca: An instruction-following llama model},
  author={Taori, Rohan and Gulrajani, Ishaan and Zhang, Tianyi and Dubois, Yann and Li, Xuechen and Guestrin, Carlos and Liang, Percy and Hashimoto, Tatsunori B},
  year={2023},
  publisher={Stanford, CA, USA}
}

@article{zheng2023judging,
  title={Judging llm-as-a-judge with mt-bench and chatbot arena},
  author={Zheng, Lianmin and Chiang, Wei-Lin and Sheng, Ying and Zhuang, Siyuan and Wu, Zhanghao and Zhuang, Yonghao and Lin, Zi and Li, Zhuohan and Li, Dacheng and Xing, Eric and others},
  journal={Advances in neural information processing systems},
  volume={36},
  pages={46595--46623},
  year={2023}
}

@article{cobbe2021training,
  title={Training verifiers to solve math word problems},
  author={Cobbe, Karl and Kosaraju, Vineet and Bavarian, Mohammad and Chen, Mark and Jun, Heewoo and Kaiser, Lukasz and Plappert, Matthias and Tworek, Jerry and Hilton, Jacob and Nakano, Reiichiro and others},
  journal={arXiv preprint arXiv:2110.14168},
  year={2021}
}

@book{saad2003iterative,
  title={Iterative methods for sparse linear systems},
  author={Saad, Yousef},
  year={2003},
  publisher={SIAM}
}

@inproceedings{ainslie2023gqa,
  title={Gqa: Training generalized multi-query transformer models from multi-head checkpoints},
  author={Ainslie, Joshua and Lee-Thorp, James and De Jong, Michiel and Zemlyanskiy, Yury and Lebr{\'o}n, Federico and Sanghai, Sumit},
  booktitle={Proceedings of the 2023 conference on empirical methods in natural language processing},
  pages={4895--4901},
  year={2023}
}

@article{zhang2019root,
  title={Root mean square layer normalization},
  author={Zhang, Biao and Sennrich, Rico},
  journal={Advances in neural information processing systems},
  volume={32},
  year={2019}
}

@article{shazeer2020glu,
  title={Glu variants improve transformer},
  author={Shazeer, Noam},
  journal={arXiv preprint arXiv:2002.05202},
  year={2020}
}

@article{su2024roformer,
  title={Roformer: Enhanced transformer with rotary position embedding},
  author={Su, Jianlin and Ahmed, Murtadha and Lu, Yu and Pan, Shengfeng and Bo, Wen and Liu, Yunfeng},
  journal={Neurocomputing},
  volume={568},
  pages={127063},
  year={2024},
  publisher={Elsevier}
}

@article{loshchilov2017decoupled,
  title={Decoupled weight decay regularization},
  author={Loshchilov, Ilya and Hutter, Frank},
  journal={arXiv preprint arXiv:1711.05101},
  year={2017}
}

@inproceedings{narayanan2021efficient,
  title={Efficient large-scale language model training on gpu clusters using megatron-lm},
  author={Narayanan, Deepak and Shoeybi, Mohammad and Casper, Jared and LeGresley, Patrick and Patwary, Mostofa and Korthikanti, Vijay and Vainbrand, Dmitri and Kashinkunti, Prethvi and Bernauer, Julie and Catanzaro, Bryan and others},
  booktitle={Proceedings of the international conference for high performance computing, networking, storage and analysis},
  pages={1--15},
  year={2021}
}

@article{snell2024scaling,
  title={Scaling llm test-time compute optimally can be more effective than scaling model parameters},
  author={Snell, Charlie and Lee, Jaehoon and Xu, Kelvin and Kumar, Aviral},
  journal={arXiv preprint arXiv:2408.03314},
  year={2024}
}
\bibliographystyle{plainnat}

\clearpage
\appendix

\section{Model Configurations and Training Details}
\label{app:training_details}

\subsection{Model Architecture}

\paragraph{Common backbone.}
All models use a Qwen3-style decoder-only Transformer architecture~\citep{vaswani2017attention,yang2025qwen3} and are trained from random initialization.
The backbone uses grouped-query attention (GQA)~\citep{ainslie2023gqa}, Q/K normalization, pre-RMSNorm~\citep{zhang2019root}, SwiGLU activations with SiLU~\citep{shazeer2020glu}, and rotary position embeddings (RoPE)~\citep{su2024roformer}.
The attention projections do not use bias terms.
We use a vocabulary size of 50,304 and tie the input embedding and output language-model head.
Unless otherwise specified, all models use
\texttt{rms\_norm\_eps}$=10^{-6}$,
\texttt{rope\_theta}$=10^{6}$,
\texttt{attention\_dropout}$=0$,
and \texttt{initializer\_range}$=0.02$.
The attention head dimension is fixed to 64 across all model sizes.

For every model size, the Vanilla and P2N variants have exactly the same number of parameters and the same physical Transformer layers.
P2N does not introduce additional parameters.
Throughout the paper, Transformer layers are indexed from zero.
The corresponding implementation option \texttt{p2n\_core\_start\_layer} uses one-based indexing.

\subsubsection{Dense Models}

\autoref{tab:dense_configs} summarizes the dense model configurations.
As the model size increases, we scale depth, hidden dimension, and FFN width while preserving a middle-third P2N core.

\begin{table}[!htbp]
\centering
\small
\caption{
Dense model configurations.
Vanilla and P2N use identical physical architectures and parameter counts.
The P2N core is reported using zero-based layer indices.
}
\label{tab:dense_configs}
\resizebox{\textwidth}{!}{
\begin{tabular}{@{}crrrrrrcc@{}}
\toprule
Model &
Parameters &
Layers &
$d_{\mathrm{model}}$ &
$d_{\mathrm{ff}}$ &
Q heads &
KV heads &
P2N core &
Prefix/Core/Suffix \\
\midrule
150M
& 149,541,120
& 12
& 768
& 3,328
& 12
& 4
& L4--L7
& 4/4/4 \\

300M
& 303,203,200
& 15
& 1,024
& 4,608
& 16
& 4
& L5--L9
& 5/5/5 \\

600M
& 604,627,328
& 21
& 1,280
& 5,632
& 20
& 5
& L7--L13
& 7/7/7 \\

1B
& 1,011,626,496
& 24
& 1,536
& 7,168
& 24
& 6
& L8--L15
& 8/8/8 \\

1.5B
& 1,496,092,288
& 27
& 1,792
& 8,192
& 28
& 7
& L9--L17
& 9/9/9 \\
\bottomrule
\end{tabular}
}
\end{table}

\subsubsection{Mixture-of-Experts Models}

We additionally evaluate P2N on sparse Mixture-of-Experts (MoE) models constructed from the corresponding dense backbones.
Every Transformer layer contains 64 routed experts and activates the top 8 experts for each token, with no additional shared expert.
The intermediate dimension of each individual expert is set to one eighth of the FFN width of the corresponding dense model.
Consequently, the aggregate FFN width of the eight activated experts matches that of the dense counterpart.
The active parameter count is therefore closely matched to the corresponding dense scale, with a small increase primarily due to the routing module.

The MoE models retain the same attention configuration, normalization scheme, positional encoding, vocabulary, tied embeddings, and P2N core placement as their dense counterparts.
We use Qwen-style unscaled output-projection initialization for the MoE models.

\begin{table}[!htbp]
\centering
\small
\caption{
MoE model configurations.
``Active parameters'' denote the parameters activated for each token rather than the total number of model parameters.
All models use 64 routed experts with Top-8 routing.
}
\label{tab:moe_configs}
\resizebox{\textwidth}{!}{
\begin{tabular}{@{}crrrrrccc@{}}
\toprule
Model &
Active params. &
Total params. &
Layers &
$d_{\mathrm{model}}$ &
$d_{\mathrm{ff}}^{\mathrm{expert}}$ &
Q/KV heads &
P2N core &
Prefix/Core/Suffix \\
\midrule
A0.15B
& 150,130,944
& 794,218,752
& 12
& 768
& 416
& 12/4
& L4--L7
& 4/4/4 \\

A0.3B
& 304,186,240
& 1,790,542,720
& 15
& 1,024
& 576
& 16/4
& L5--L9
& 5/5/5 \\

A0.6B
& 606,347,648
& 3,785,499,008
& 21
& 1,280
& 704
& 20/5
& L7--L13
& 7/7/7 \\

A1B
& 1,013,985,792
& 6,563,049,984
& 24
& 1,536
& 896
& 24/6
& L8--L15
& 8/8/8 \\
\bottomrule
\end{tabular}
}
\end{table}

\subsection{Training Setup}
\label{app:training_setup}

All models are pretrained from scratch using the standard causal next-token prediction objective.
Training examples have a sequence length of 2,048 tokens.
We use a global batch size of 1,024 sequences, corresponding to 2,097,152 training tokens per optimizer step, with a per-device micro-batch size of 8 sequences.

For the dense models, we use AdamW~\citep{loshchilov2017decoupled}.
For the MoE models, we use the Megatron Adam optimizer with the distributed optimizer~\citep{narayanan2021efficient}.
Both settings use
$\beta_1=0.9$,
$\beta_2=0.95$,
$\epsilon=10^{-8}$,
weight decay of $0.1$,
and global gradient-norm clipping at $1.0$.

We employ a cosine learning-rate schedule with a linear warmup over the first 5\% of training, followed by cosine decay to 10\% of the peak learning rate.
For the 150M, 300M, and 600M dense models and their corresponding MoE variants, we use a peak learning rate of $1.5\times10^{-3}$.
For the 1B dense model and A1B MoE model, the peak learning rate is $1.0\times10^{-3}$, while the 1.5B dense model uses $8.0\times10^{-4}$.

Training uses BF16 computation throughout.
For the MoE experiments, the main model parameters and accumulated gradients are maintained in FP32 by the Megatron optimizer.
We use a random seed of 42 for all experiments.
Activation recomputation is disabled in the reported runs.

Unless explicitly varied in an ablation, Vanilla and P2N models at the same scale use the same architecture, global batch size, optimizer hyperparameters, learning-rate schedule, and training protocol.
Thus, the comparison changes only the P2N computation defined in~\autoref{sec:method}, while keeping the underlying parameterization and optimization setup fixed.

\section{Supervised Fine-Tuning Details}
\label{app:sft_details}

We conduct supervised fine-tuning (SFT) for two settings: mathematical reasoning on GSM8K and instruction tuning on Alpaca followed by evaluation on MT-Bench.
For each setting, the Vanilla and P2N models are independently initialized from their corresponding pretrained checkpoints and fine-tuned with identical data and optimization hyperparameters.
All experiments use full-parameter SFT.
For P2N, we retain the same architecture and training recurrence as in pretraining, with the number of Jacobi updates randomly sampled from $K\in\{2,3\}$ for each training batch.

\subsection{GSM8K Supervised Fine-Tuning}
\label{app:gsm8k_sft}

For mathematical reasoning, we fine-tune the models on the official GSM8K training split, which contains 7,473 examples.
The official test split of 1,319 examples is reserved exclusively for evaluation and is not used during SFT.
Each example places the problem after the label ``Question:''
and starts the next line with ``Answer:''; the original GSM8K
response serves as the target.
The target retains both the intermediate reasoning trace and the final answer indicated by the \texttt{\#\#\#\#} marker.
Only response tokens contribute to the training objective; prompt tokens are masked from the loss.

Both Vanilla and P2N models are trained for 3 epochs with a global batch size of 128 and a maximum sequence length of 2,048 tokens.
We do not use sequence packing.
Optimization uses AdamW with a peak learning rate of $2\times10^{-5}$,
$\beta_1=0.9$,
$\beta_2=0.95$,
$\epsilon=10^{-8}$,
zero weight decay,
and gradient clipping with a maximum global norm of $1.0$.
The learning rate follows a cosine schedule with a warmup ratio of $0.03$.
Training is performed in BF16 with a random seed of 42.

\subsection{Alpaca Supervised Fine-Tuning and MT-Bench Evaluation}
\label{app:alpaca_mtbench_sft}

For instruction-following evaluation, we first perform SFT on the English Alpaca dataset and subsequently evaluate the resulting models on MT-Bench.
MT-Bench itself is used only for evaluation and is not included in the training data.

We use a fixed snapshot of the \texttt{llamafactory/alpaca\_en} dataset containing 51,699 training examples.
We adopt the standard instruction-tuning format and optimize only over response tokens, with the instruction and input prompt masked from the loss.
Models are trained for 3 epochs with a global batch size of 128, a maximum sequence length of 2,048 tokens, and sequence packing disabled.

We use AdamW with a peak learning rate of $2\times10^{-5}$,
$\beta_1=0.9$,
$\beta_2=0.95$,
$\epsilon=10^{-8}$,
zero weight decay,
and a maximum gradient norm of $1.0$.
The learning rate follows a cosine schedule with a warmup ratio of $0.03$.
Training uses BF16 precision and a random seed of 42.
As with GSM8K, Vanilla and P2N models use exactly the same SFT data and optimization settings.

After Alpaca SFT, we evaluate instruction-following and multi-turn conversational ability on MT-Bench.
Models generate responses to the standard MT-Bench questions under the same generation protocol, and the resulting answers are evaluated by \texttt{GPT-5.5} as the judge model.
The judge is used only for post-training evaluation and does not participate in SFT.

\section{Additional Experiments on Mixture-of-Experts Models}
\label{app:moe_experiments}

We extend our experiments to sparse Mixture-of-Experts (MoE)
models to examine whether the benefits of \method{} generalize
beyond dense Transformers.
Following the organization of~\autoref{sec:experiments},
we evaluate scaling behavior, language modeling, downstream
performance, decoding efficiency, and the effect of additional
Jacobi iterations.

\subsection{Large-Scale Pretraining}
\label{app:moe_pretraining}

We pretrain Qwen-style MoE models from scratch on the Pile,
covering active parameter scales of A0.15B, A0.3B, A0.6B,
and A1B.
The prefix ``A'' denotes the number of parameters activated
per token rather than the total model parameter count.
Each Transformer layer contains 64 routed experts and activates
the top 8 experts per token, without an additional shared expert.
For each scale, Vanilla and \method{} share the same backbone
architecture, active and total parameter counts, training data,
token budget, and optimization settings.
As in the dense experiments, \method{} applies feedback to
the middle third of the network and samples
$K\in\{2,3\}$ Jacobi updates during training.
Detailed model configurations and training hyperparameters
are provided in~\autoref{app:training_details}.

\subsubsection{Scaling Behavior}
\label{app:moe_scaling}

\begin{figure}[!htbp]
    \centering
    \includegraphics[width=\linewidth]{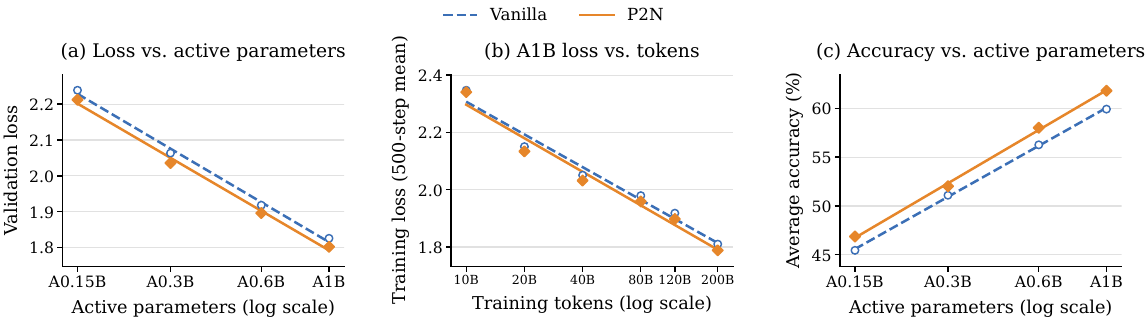}
    \caption{\textbf{Scaling behavior of MoE models.}
    \textbf{(a)} Final Pile validation loss across active
    parameter scales, from A0.15B to A1B.
    \textbf{(b)} Training loss over training tokens for the
    matched A1B models, with each point averaged over
    500 consecutive optimizer steps.
    \textbf{(c)} Mean raw accuracy over all nine downstream
    task configurations across active parameter scales.
    Lower is better in panels~(a) and~(b), and higher is better
    in panel~(c).
    \method{} achieves lower final validation loss, lower
    training loss at the reported token budgets, and higher
    average downstream accuracy.}
    \label{fig:moe-scaling}
\end{figure}

\autoref{fig:moe-scaling} summarizes the scaling behavior
of the MoE models.
Panels~(a) and~(c) report final Pile validation loss and average
downstream accuracy across active parameter scales, respectively,
while panel~(b) tracks training loss for the matched A1B models.
The downstream score is the unweighted mean of raw accuracies, using the same aggregation
as~\autoref{tab:moe-downstream}.
All horizontal axes use a logarithmic scale.

Across all four active parameter scales, \method{} achieves
lower validation loss and higher average downstream accuracy
than Vanilla.
The validation-loss reductions range from 0.0219 to 0.0275,
while the accuracy improvements range from 0.92 to 1.88
percentage points.
At A1B, final validation loss decreases from 1.8259 to 1.8021,
and average downstream accuracy increases from 59.92\%
to 61.80\%.
For the matched A1B training runs, \method{} also maintains
lower training loss at every reported token budget from
10B to 200B tokens.
These results demonstrate consistent improvements across
active parameter scales and the sampled stages of pretraining.

\subsubsection{Language Modeling Ability}
\label{app:moe_language_modeling}

\begin{figure}[!htbp]
    \centering
    \includegraphics[width=\linewidth]{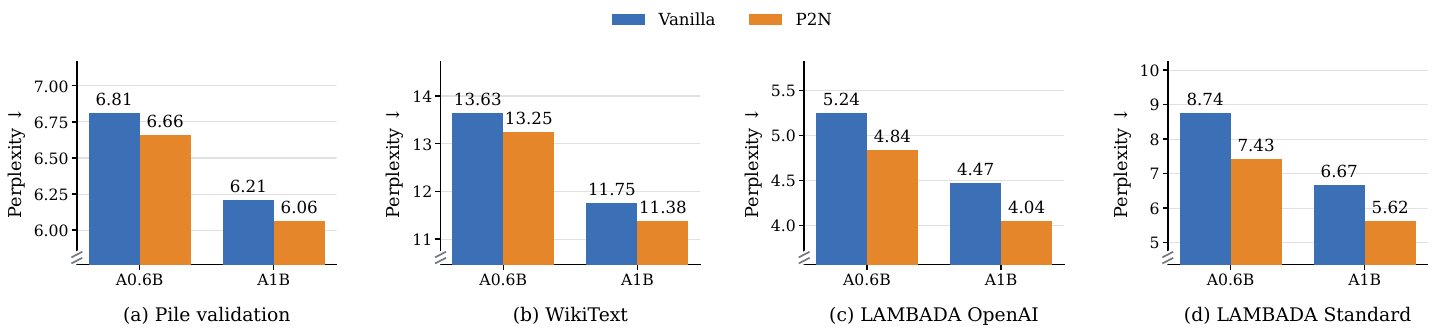}
    \caption{\textbf{Language modeling at the A0.6B and A1B scales.}
    Perplexity of Vanilla and \method{} MoE models on
    \textbf{(a)} Pile validation,
    \textbf{(b)} WikiText,
    \textbf{(c)} LAMBADA OpenAI, and
    \textbf{(d)} LAMBADA Standard; lower is better.
    Numbers above the bars denote perplexity values.
    \method{} achieves lower perplexity at both scales on
    every benchmark, with the largest relative improvements
    on the LAMBADA variants.}
    \label{fig:moe-language-modeling}
\end{figure}

We evaluate the A0.6B and A1B MoE models on Pile validation,
WikiText, and the OpenAI and Standard variants of LAMBADA.
As shown in ~\autoref{fig:moe-language-modeling},
\method{} consistently reduces perplexity on all four
benchmarks at both active parameter scales.

Relative to Vanilla, the perplexity reductions are
2.2--2.4\% on Pile,
2.8--3.2\% on WikiText,
7.7--9.4\% on LAMBADA OpenAI,
and 15.0--15.7\% on LAMBADA Standard.
At A1B, for example, Pile perplexity decreases from 6.2081
to 6.0621, while LAMBADA Standard perplexity decreases from
6.6722 to 5.6217.
The largest relative improvements occur on the two LAMBADA
variants, consistent with the pattern observed in the dense
experiments.
These results show that the language-modeling gains extend
to sparse MoE architectures.

\subsection{Downstream Task Evaluation}
\label{app:moe_downstream}

We further assess whether the improvements observed during
MoE pretraining translate into better downstream performance.
We evaluate general downstream tasks at the A0.6B and A1B
scales and mathematical reasoning at the A1B scale.

\subsubsection{General Downstream Tasks}
\label{app:moe_general_downstream}

We evaluate the MoE models on the same nine task configurations
as in ~\autoref{sec:general-downstream}:
LAMBADA-OpenAI, LAMBADA-Standard, PIQA, WinoGrande, SciQ,
HellaSwag, ARC-Easy, ARC-Challenge, and RACE.
~\autoref{tab:moe-downstream} reports raw accuracy and the
unweighted average over all nine configurations.

At A0.6B, \method{} improves average accuracy from 56.26\%
to 58.00\%, a gain of 1.75 percentage points, with improvements
on all nine task configurations.
At A1B, average accuracy increases from 59.92\% to 61.80\%,
a gain of 1.88 percentage points.
The A1B model improves on eight configurations and matches
Vanilla on RACE.
The largest absolute gains at both scales occur on
LAMBADA-Standard, reaching 4.50 and 4.37 percentage points
at A0.6B and A1B, respectively.
These results demonstrate that the benefits of \method{}
extend to general downstream performance in MoE models.

\begin{table}[!htbp]
    \centering
    \caption{\textbf{Downstream performance at the A0.6B and A1B scales.}
    Raw accuracy (\%) of Vanilla and \method{} MoE models.
    Avg. denotes the unweighted mean over the nine task
    configurations.
    Bold indicates the best score within each matched scale,
    including ties.
    \method{} improves average accuracy at both scales.}
    \label{tab:moe-downstream}

    \footnotesize
    \setlength{\tabcolsep}{2.5pt}
    \renewcommand{\arraystretch}{1.18}

    \begin{tabularx}{\linewidth}{
        @{}
        c
        l
        *{10}{>{\centering\arraybackslash}X}
        @{}
    }
        \toprule
        \multirow{2}{*}{Scale}
        & \multirow{2}{*}{Method}
        & \multicolumn{2}{c}{LAMBADA}
        & \multirow{2}{*}{PIQA}
        & \multirow{2}{*}{\shortstack{Wino\\Grande}}
        & \multirow{2}{*}{SciQ}
        & \multirow{2}{*}{\shortstack{Hella\\Swag}}
        & \multicolumn{2}{c}{ARC}
        & \multirow{2}{*}{RACE}
        & \multirow{2}{*}{Avg.} \\

        \cmidrule(lr){3-4}
        \cmidrule(lr){9-10}

        & & OpenAI & Standard
        & & & &
        & Easy & Challenge
        & & \\

        \midrule

        A0.6B & Vanilla
        & 64.43
        & 52.61
        & 72.85
        & 59.04
        & 88.60
        & 44.12
        & 63.05
        & 26.79
        & 34.83
        & 56.26 \\

        A0.6B & \method{}
        & \textbf{66.68}
        & \textbf{57.11}
        & \textbf{73.18}
        & \textbf{61.25}
        & \textbf{90.60}
        & \textbf{45.08}
        & \textbf{63.51}
        & \textbf{29.69}
        & \textbf{34.93}
        & \textbf{58.00} \\

        \midrule

        A1B & Vanilla
        & 67.77
        & 58.74
        & 75.03
        & 62.59
        & 90.20
        & 48.44
        & 68.06
        & 32.94
        & \textbf{35.50}
        & 59.92 \\

        A1B & \method{}
        & \textbf{69.84}
        & \textbf{63.11}
        & \textbf{75.79}
        & \textbf{65.35}
        & \textbf{92.30}
        & \textbf{49.62}
        & \textbf{69.70}
        & \textbf{34.98}
        & \textbf{35.50}
        & \textbf{61.80} \\

        \bottomrule
    \end{tabularx}
\end{table}

\subsubsection{Mathematical Reasoning}
\label{app:moe_mathematical_reasoning}

We fine-tune the pretrained A1B MoE models on GSM8K following
the SFT settings in~\autoref{app:sft_details}.
We evaluate strict final-answer exact match (EM) on all
1,319 test examples under zero-shot and eight-shot prompting.
Fine-tuning settings are matched between Vanilla and
\method{}, as are the prompting formats, decoding settings,
and answer-extraction procedures within each evaluation setting.

As shown in~\autoref{tab:moe-gsm8k}, \method{} improves
zero-shot EM from 21.68\% to 27.98\%, a gain of
6.29 percentage points.
Under eight-shot prompting, EM increases from 20.02\%
to 26.38\%, a gain of 6.37 percentage points.
These results demonstrate improved mathematical reasoning
performance under both prompting settings.

\begin{table}[!htbp]
    \centering
    \caption{\textbf{Mathematical reasoning at the A1B MoE scale.}
    GSM8K test strict final-answer EM (\%) under zero-shot
    and eight-shot prompting after supervised fine-tuning.
    Gains are measured in percentage points relative to
    Vanilla and calculated before rounding.
    \method{} improves accuracy in both settings.}
    \label{tab:moe-gsm8k}

    \small
    \setlength{\tabcolsep}{7pt}
    \renewcommand{\arraystretch}{1.12}

    \begin{tabular}{@{}clrrrr@{}}
        \toprule
        \multirow{2}{*}{Scale}
        & \multirow{2}{*}{Method}
        & \multicolumn{2}{c}{Zero-shot}
        & \multicolumn{2}{c}{Eight-shot} \\

        \cmidrule(lr){3-4}
        \cmidrule(lr){5-6}

        & & EM (\%) $\uparrow$ & Gain (pp.)
          & EM (\%) $\uparrow$ & Gain (pp.) \\

        \midrule

        A1B & Vanilla
        & 21.68 & --
        & 20.02 & -- \\

        A1B & \method{}
        & \textbf{27.98} & $+6.29$
        & \textbf{26.38} & $+6.37$ \\

        \bottomrule
    \end{tabular}
\end{table}

\subsection{Decoding Efficiency}
\label{app:moe_decoding_efficiency}

During autoregressive decoding, \method{} reuses the previous
token's core output through a vector addition, without repeating
any Transformer layer.
We benchmark the A1B MoE models on a single GPU
using BF16, SDPA, and the standard KV cache.
Each run uses a 256-token prompt and generates 2,048 tokens
per sequence.
For each configuration, we perform one untimed warm-up run
and report the mean over four subsequent timed runs.
Prompt prefill is excluded from the decoding measurements.

\autoref{tab:moe-decoding-efficiency} reports time per output
token (TPOT), aggregate decoding throughput, and relative latency.
Across batch sizes from 1 to 32, \method{} maintains nearly identical
TPOT and throughput to Vanilla, with relative latency remaining within
0.1\% of the baseline under all reported settings.
These results show that the parameter-free feedback introduces
negligible decoding overhead for the A1B MoE model.

\begin{table}[t]
\centering
\caption{\textbf{Autoregressive decoding efficiency of the A1B MoE model on a single GPU.}
Each entry is the mean of four timed runs generating 2,048 tokens per sequence
after one untimed warm-up run.
Each run uses a 256-token prompt.
Relative latency is normalized to Vanilla TPOT.}
\label{tab:moe-decoding-efficiency}
\small
\begin{tabular}{rccccc}
\toprule
Batch & \multicolumn{2}{c}{TPOT (ms)} &
\multicolumn{2}{c}{Throughput (tok/s)} &
Rel. latency \\
\cmidrule(lr){2-3}\cmidrule(lr){4-5}
size & Vanilla & \method{} & Vanilla & \method{} &
\method{} / Vanilla \\
\midrule
1  & 70.454 & 70.470 &  14.2 &  14.2 & 1.000$\times$ \\
4  & 70.537 & 70.464 &  56.7 &  56.8 & 0.999$\times$ \\
16 & 71.038 & 71.013 & 225.2 & 225.3 & 1.000$\times$ \\
32 & 70.905 & 70.923 & 451.3 & 451.2 & 1.000$\times$ \\
\bottomrule
\end{tabular}
\end{table}

\subsection{Convergence Diagnostics of Jacobi Iteration}
\label{app:moe_jacobi_diagnostics}

\begin{figure}[!htbp]
    \centering
    \includegraphics[width=\linewidth]{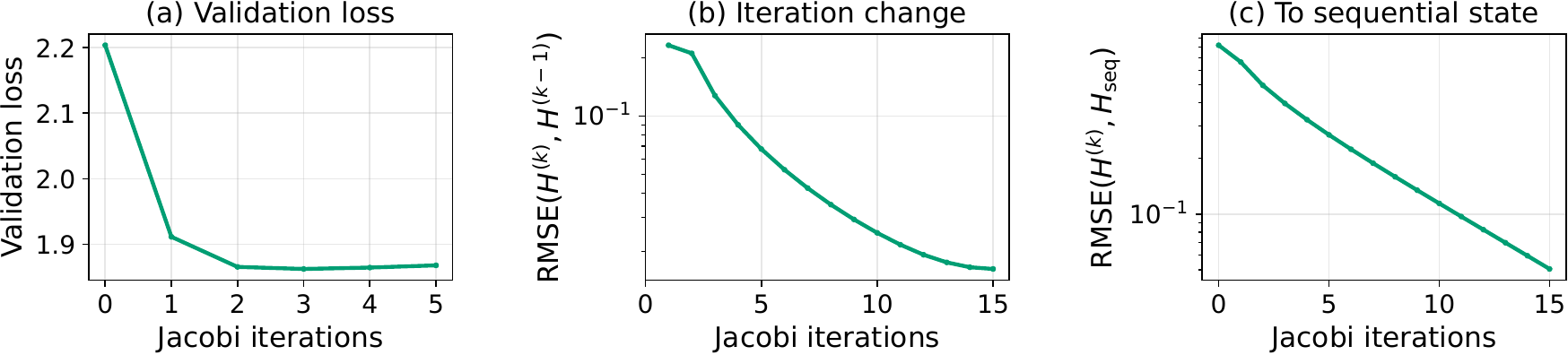}
    \caption{\textbf{Evaluation loss under additional Jacobi
    iterations at the A1B MoE scale.}
    Loss is evaluated on 4,096 examples for $K=0,\ldots,5$,
    where $K$ counts updates after the initial warm pass.
    After the initial reduction, evaluation loss changes
    only slightly as further iterations are applied.}
    \label{fig:moe-jacobi-diagnostics}
\end{figure}

We examine evaluation loss as a function of the number of
Jacobi iterations on the A1B \method{} MoE model.
As shown in~\autoref{fig:moe-jacobi-diagnostics}, loss
decreases over the first few updates and changes only slightly
thereafter.
In particular, it decreases from 2.2034 at $K=0$ to 1.8656
at $K=2$ and remains between 1.8625 and 1.8680 for
$K=2,\ldots,5$.
Thus, additional iterations beyond the training range produce
only small changes in evaluation loss over the tested range.

\section{Limitations and Future Work}
\label{app:limitations}

Jacobi iteration preserves parallel training but increases the
estimated training compute to approximately $1.83\times$ that
of Vanilla under our default configuration and layer-based
compute accounting.
In addition, the dependence on the previous token's hidden state
means that efficient integration with multi-token prediction
and speculative decoding requires further investigation.
We are actively exploring more efficient training schedules
and compatible decoding strategies to support practical deployment.

\end{document}